\documentclass[sn-mathphys,Numbered]{sn-jnl}

\usepackage{graphicx}%
\usepackage{multirow}%
\usepackage{amsmath,amssymb,amsfonts}%
\usepackage{amsthm}%
\usepackage{mathrsfs}%
\usepackage[title]{appendix}%
\usepackage{xcolor}%
\usepackage{textcomp}%
\usepackage{manyfoot}%
\usepackage{booktabs}%
\usepackage{algorithm}%
\usepackage{algorithmicx}%
\usepackage{algpseudocode}%
\usepackage{listings}%
\usepackage[vietnamese,english]{babel}
\usepackage[T5,T1]{fontenc}

\newcommand*\vn{\fontencoding{T5}\selectfont\selectlanguage{vietnamese}}
\newcommand*\en{\fontencoding{T1}\selectfont\selectlanguage{english}}
\theoremstyle{thmstyleone}%
\theoremstyle{thmstyletwo}%

\theoremstyle{thmstylethree}%

\begin{document}

\title[Vietname GPT for COVID-19]{Generative Pre-trained Transformer for Vietnamese Community-based COVID-19 Question Answering}

\author[1,2]{ \sur{Tam Minh Vo}}\email{tamvm@uit.edu.vn}

\author*[1,2]{\sur{Tran Vinh Khiem}}\email{khiemtv@uit.edu.vn}

\affil[1]{\orgname{University of Information Technology}, \orgaddress{ \city{Ho Chi Minh city}, \country{Vietnam}}}

\affil[2]{\orgname{Vietnam National University}, \orgaddress{ \city{Ho Chi Minh city}, \country{Vietnam}}}


\abstract{Recent studies have provided empirical evidence of the wide-ranging potential of Generative Pre-trained Transformer (GPT), a pretrained language model, in the field of natural language processing. GPT has been effectively employed as a decoder within state-of-the-art (SOTA) question answering systems, yielding exceptional performance across various tasks. However, the current research landscape concerning GPT's application in Vietnamese remains limited. This paper aims to address this gap by presenting an implementation of GPT-2 for community-based question answering specifically focused on COVID-19 related queries in Vietnamese. We introduce a novel approach by conducting a comparative analysis of different Transformers vs SOTA models in the community-based COVID-19 question answering dataset. The experimental findings demonstrate that the GPT-2 models exhibit highly promising outcomes, outperforming other SOTA models as well as previous community-based COVID-19 question answering models developed for Vietnames.}


\maketitle

\section{Introduction}\label{sec1}

Community-based Question Answering (CQA) is a prominent task that involves soliciting
answers from the collective intelligence of online communities \cite{Zhang_Chen_Dong_Wang_Zha_Wang_2021}. Users seeking information
post their queries on publicly accessible websites or forums, where others contribute their
responses. This interactive question-answering approach has gained popularity in everyday
life, with platforms like Quora and Reddit serving as prominent examples. CQA platforms
facilitate the seamless exchange of questions and answers, enabling users to tap into the
collective knowledge of the community. The advancement of question answering systems has
empowered computers to comprehend and respond to user queries.

During the period from 2020 to 2021, Vietnam, like many other countries, faced the challenges
posed by the COVID-19 pandemic. Access to accurate information regarding the virus,
preventive measures, self-quarantine guidelines, vaccination policies, and government
regulations aimed at curbing the transmission of the virus became crucial for citizens.
Consequently, individuals frequently pose inquiries concerning the COVID-19 situation,
appropriate actions when exposed to COVID-19 patients, vaccination policies, and other
relevant topics.

In the light of these circumstances, our motivation for conducting this study is to facilitate the
evaluation of a COVID-19-focused question answering system based on Community Question
Answering (CQA) in the Vietnamese language. We aim to investigate the disparities observed
among various pretrained transformer-based language models when fine-tuned for the task of
question answering, particularly in adapting to datasets with varying levels of complexity.
Furthermore, we conduct a comparative analysis between these models and existing state-ofthe-art techniques using a community-based dataset specifically focused on COVID-19- related
questions. The study encompasses an examination of three approaches, namely traditional neural network, traditional transformer, and generative pretrained transformer. Given the
prominence of ChatGPT and the widespread attention it has received, our research seeks to
explore the potential and capabilities of the Generative Pre-Trained Transformer (GPT) model
in the Vietnamese language context.

The subsequent sections of this paper are organized as follows. Section 2 provides an overview
of relevant literatures in the field. In Section 3, comprehensive descriptions of our
methodologies: the Convolutional Neural Network architectures, Transformer and the
Generative Pre-trained Transformer (GPT) are presented, respectively. The dataset and
outcomes of our experimental evaluations, along with the benchmarks and error analysis, are
outlined in Section 4. Finally, Section 5 encompasses our concluding remarks andd highlights
potential directions for future research.
\section{Related work}
In contrast, several efforts have been dedicated to the development of question answering
corpora specifically designed for the Vietnamese language. Nguyen introduced UITViQuAD
\cite{nguyen-etal-2020-vietnamese}, UIT-ViNewsQA \cite{10.1145/3527631}, and ViMMRC \cite{9247161}, all these datasets focus on text-based question
answering tasks. Khanh, Nghia and Khiem proposed ViVQA \cite{tran-etal-2021-vivqa-vietnamese}, OpenViVQA \cite{NGUYEN2023101868} and ViCLEVR \cite{tran2023viclevr} as a dataset specifically designed for visual-based
question answering, while Luu introduced ViCoQA \cite{10.1007/978-3-030-88113-9_44} for conversational question answering
task. Notably, Thai presented UIT-ViCoV19QA \cite{thai-etal-2022-uit} as the pioneering dataset that encompasses
COVID-19-related information gathered from the community. According to our knowledge, this
is also a rare dataset for the generative question answering task for Vietnamese.

Regarding question answering datasets related to COVID-19, Tran (2022) conducted a
comparative study with various knowledge-based question answering methods and proposed a
new bilingual dataset, COVID-KGQA \cite{10.1007/978-3-031-22064-7_20}, which includes Vietnamese as one of the languages.
However, the authors primarily focused on English and did not specifically emphasize the
Vietnamese language.

In this work, we aim to conduct a comparative study of generative question answering task,
specifically focusing on the topic of COVID-19 with a primary emphasis on Vietnamese.
Therefore, we decided to choose the UIT-ViCoV19QA \cite{thai-etal-2022-uit} dataset for our experiments, as well
as inherit the authors' previous experiments in addition to proposing a new method for
comparison purposes.
\section{Methodologies}
\subsection{Neural Network}
\subsubsection{Attention-based Recurrent Neural Network}
The experimental recurrent neural network (RNN) models employed in this study utilize two
attention mechanisms: Bahdanau Attention \cite{DBLP:journals/corr/BahdanauCB14} and Luong Attention \cite{luong-etal-2015-effective}. To ensure
comparability of the results, both models are configured with similar hyperparameters in both
the encoder and decoder components.

Specifically, an embedding layer with a dimension of 512 is utilized to represent the input text
in both RNN-1 and RNN-2. The hidden layers consist of two gated recurrent unit (GRU) cells,
each with a hidden size of 512. A dropout rate of 0.5 is applied to mitigate overfitting and
improve generalization.

To enhance the contextual understanding of the models, a bidirectional gated recurrent unit
(BiGRU) is implemented in the encoder component of both RNN-1 and RNN-2. This allows
the models to capture information from both the forward and backward sequences, thereby
improving their ability to comprehend the context. 

For ease of reference, the RNN model incorporating Bahdanau attention is referred to as RNN1, while the model incorporating Luong attention is referred to as RNN-2. By employing these
specific hyperparameters and attention mechanisms, we aim to evaluate and compare the
performance of the RNN models in the task at hand.
\subsubsection{Convolutional Neural Network}
In contrast to recurrent neural networks, convolutional neural networks (CNNs) \cite{DBLP:conf/ijcai/QiuH15} leverage
multiple convolutional layers that are commonly used in image processing. These layers are
equipped with filters that aid in extracting distinct features from textual data. In our
experimental investigation, we have established specific hyperparameters for the CNN models.
The CNN models utilized an embedding layer with a dimensionality of 512 to represent the
input text. Additionally, three convolutional layers were employed, each with a hidden size of
512. These layers utilized 1024 filters with a kernel size of 3 x 3, which allowed them to capture
different patterns and features within the text. Furthermore, a drop-out probability of 0.5 was
applied to mitigate overfitting and enhance generalization performance.
By employing these hyperparameters, we aimed to optimize the performance of the CNN
models in our experimental setup and enable them to effectively extract meaningful features
from textual data.
\subsubsection{Transformer}
In this study, we have utilized the transformer model proposed in the paper “Attention is all
you need” \cite{DBLP:conf/ijcai/QiuH15}. The transformer model has gained significant prominence in the field of natural
language processing (NLP) and has been widely adopted for achieving state-of-the-art
performance across various tasks. Additionally, derivative versions of the transformer model,
such as BERT \cite{devlin-etal-2019-bert}(Bidirectional Encoder Representations from Transformers) and its pretrained adaptations, have also demonstrated exceptional performance.

In our study, we have configured the transformer model with specific specifications. These
include an embedding layer with a dimension of 512, two transformer layers that incorporate
8 self-attention heads, a positional embedding layer with a maximum sequence length of 500,
a position-wise feed-forward layer with a dimension of 2048, and a dropout rate of 0.5. These
specifications have been chosen based on prior research and empirical evidence to optimize the
performance of the transformer model for the specific task at hand.
\subsubsection{Generative Pre-trained Transformer}
Generative Pre-trained Transformer (GPT-2) \cite{radford2019language}, a transformer-based model, has undergone a
comprehensive pretraining process on a large corpus of English text data. This pre-training
process is characterized by self-supervision, eliminating the need for human annotation or
labeling. Instead, GPT-2 utilizes publicly available texts to automatically generate inputs and
corresponding labels. The primary objective of GPT-2's pretraining is to predict the next word
in a given sentence, enabling it to learn the intricacies of language.

With a substantial parameter count of 1.5 billion, GPT-2 ranks among the largest language
models developed to date. By leveraging its predictive capabilities to determine the subsequent
word in a word sequence, GPT-2 demonstrates the ability to generate text that closely
resembles human-generated content. This extensive training regimen equips the model with
the capacity to capture and replicate complex patterns, making it highly proficient in various
natural language processing tasks. The robustness and versatility of GPT-2 position it as a
powerful tool for tackling a wide array of linguistic challenges.

In addition to evaluating the performance of the original GPT model, this study also
incorporates the use of a pre-trained GPT-2 model specifically designed for the Vietnamese
language. The purpose of this comparison is to assess the performance and effectiveness of the
GPT-2 model when applied to the Vietnamese context.

\section{Experiments}\label{sec2}

\subsection{Dataset}
UIT-ViCoV19QA \cite{thai-etal-2022-uit} is an innovative and comprehensive community-driven generative
question answering dataset specifically focused on the COVID-19 pandemic in the Vietnamese 
5
context. The dataset consists of 4,500 question-answer pairs, ensuring each question has at least
one answer and can have up to four distinct paraphrased answers to enhance diversity.
The question-answer pairs were meticulously selected from publicly available FAQ documents
provided by reputable healthcare organizations such as the CDC, UNICEF, the Ministry of
Health of Vietnam, Vietnam Government Portal, and other trusted medical institutions. The
dataset covers a wide range of crucial topics related to COVID-19, including the disease's
origin, outbreak, and nomenclature, as well as its spread, symptoms, prevention measures,
treatment guidelines, nutrition, treatment models, COVID-19 variants, vaccines and
vaccination, travel regulations, isolation and quarantine protocols, lockdown and social
distancing measures, policies and sanctions, financial support, post-COVID-19 considerations,
and specific aspects concerning COVID-19 in children. The incorporation of these topics
ensures the dataset's richness and relevance in addressing various aspects of the pandemic.
Table 1 shows some examples of question-answer pairs in UIT-ViCoV19QA.
\begin{table}[ht]
 \caption{Several examples of question-answer pairs in UIT-ViCoV19QA dataset}
    \centering
    \begin{tabular}{p{0.35\linewidth} | p{0.6\linewidth}}
    \hline
    
    \textbf{Question } & \textbf{Answer} \\ \hline
    \vn  CDC có những khuyến cáo gì về việc làm sạch và khử trùng bề mặt? \en (English:
What recommendations does the CDC have for cleaning and disinfecting surfaces?)& \vn Làm sạch bề mặt và các đồ vật định kỳ nếu thường xuyên tiếp xúc (ví dụ hàng
ngày hay sau khi sử dụng) hoặc sau khi có khách đến nhà. Chú ý vào các bề mặt và vật thể
thường xuyên chạm vào (điện thoại, tay nắm cửa, công tắc đèn, mặt bàn, điều khiển từ xa),
vệ sinh chúng bằng chất tẩy rửa. Thực hành nghiêm các quy trình kiểm soát lây nhiễm nếu
địa phương của bạn có phát hiện ca bệnh. Thực hiện theo khuyến cáo của CDC về vệ sinh
dành cho cộng đồng. \en (English: Clean surfaces and objects periodically if frequently touched
(eg daily or after use) or after guests arrive. Pay close attention to frequently touched surfaces
and objects (phones, doorknobs, light switches, countertops, remote controls), and clean them
with detergent. Strictly practice infection control procedures if a case is detected in your
locality. Follow CDC recommendations for community hygiene.) \\ \hline
  \vn  Người quản lý phải làm gì nếu nhân viên hoặc cư dân xuất hiện các triệu chứng
bệnh COVID-19? \en (English: What should a manager do if residents or employees display
symptoms of COVID-19?) & \vn Yêu cầu cư dân tích cực theo dõi (ít nhất là hàng ngày) các triệu chứng bệnh
COVID-19, bao gồm sốt và các triệu chứng hô hấp (hụt hơi, khởi phát hoặc ho có thay đổi).
Người quản lý phải phối hợp với cơ quan y tế địa phương để thiết lập quy trình cho người bị
bệnh. Bệnh nhân phải tránh tiếp xúc với người khỏe mạnh. Nếu bạn cho rằng ai đó có thể đã
nhiễm COVID-19, hãy yêu cầu họ tự cách ly và bạn cần liên hệ với nhân viên y tế địa phương
ngay lập tức. \en (English: Ask residents to actively monitor (at least daily) for symptoms of
COVID-19 illness, including fever and respiratory symptoms (shortness of breath, new onset
or changing cough). Managers must coordinate with local health authorities to establish
procedures for sick people. Patients must avoid contact with healthy people. If you think
someone may have COVID-19, ask them to self-isolate and you should contact local health
officials immediately.)
\\ \hline
\vn Làm cách nào để tôi biết vi-rút đã lây lan đến cộng đồng địa phương gần nơi ở
của tôi? \en (English: How do I know if the virus has spread to the local community near me?) & \vn Chào bạn. Bạn có thể nhận các thông tin về hoạt động chống dịch COVID-19 ở
tại địa phương của mình bằng cách liên lạc với nhân viên y tế hoặc theo dõi các trang thông
tin của cơ sở y tế địa phương, đồng thời thường xuyên cập nhật thông tin trên trang web của
CDC. \en (English: Hello. You can receive information about COVID-19 anti-pandemic
activities in your locality by contacting medical staff or following the information pages of
local medical facilities, and regularly update information on the CDC website.)
\\
\hline

    \end{tabular}
    \label{tab:my_label}
\end{table}

\subsection{Evaluation Metric}
\subsubsection{BLEU}
BLEU \cite{papineni-etal-2002-bleu} is a commonly employed evaluation metric in Machine Translation (MT) that relies
on n-gram analysis. It aims to establish a strong correlation with human evaluations of
translation quality. The metric calculates the overlap of n-grams between the target text
(reference) and the predicted text (candidate) by considering the maximum count of each ngram and restricting the count of n-grams in the candidate to the maximum count found in the
reference. In our experiments, we compute the BLEU score using unigram analysis (BLEU-1)
as well as cumulative 4-gram analysis (BLEU-4).
\subsubsection{METOR}
METOR \cite{banerjee-lavie-2005-meteor} metric utilizes the harmonic mean of unigram precision and recall, giving higher
weightage to recall compared to precision. Additionally, METOR incorporates several unique
features that are absent in other metrics, including stemming and synonymy matching, in
addition to the conventional exact word matching. However, it is worth nothing that these
features are currently not available for the Vietnamese language.
\subsubsection{ROGUE-L}
The ROGUE-L \cite{lin-2004-rouge} metric evaluates the longest common subsequence (LCS) between the
output of our model and the reference answer. The basic concept is that a longer shared
sequence indicates a higher degree of similarity between the two texts. The main advantage of
using LCS is the ability to account for non-consecutive matches while still adhering to the
sentence-level word order. For each generated answer, we select the highest score attained by
comparing it against all available reference answers.
\subsection{Experiment setup}
We used the dataset UIT-ViCoV19QA, inherited the codes and settings from this repo\footnote{ https://github.com/triet2397/UIT-ViCoV19QA} by Thai
to reproduce the results of the first two model groups for comparison purposes. For the
GPT-2 architecture, we use three sets of pretrained models, which are the multilingual
pretrained set provided by its authors and two pretrained sets trained on the Vietnamese
datasets of tuanle\footnote{ https://huggingface.co/tuanle/VN-News-GPT2/}
and danghuy19993\footnote{ https://huggingface.co/danghuy1999/gpt2-viwiki/}
. All of them are provided by HuggingFace\footnote{https://huggingface.co/}
library.

Our research is conducted using a GPU Quadro RTX A6000 with 24GB VRAM and a CPU
Intel Core i9-10900X @ 3.70 GHz with 16GB RAM. These hardware specifications provide
us with the necessary computational power and resources to perform our experiments and
analyses effectively.
\subsection{Main Results}
The main findings of our comparative study highlight the performance of various neural
network approaches, including RNN-1, RNN-2, CNN, Transformer, GPT-2 original, GPT-2
with pretrained weights from tuanle and danghuy1999, in the context of the question answering
task. The overall result is shown in Table 2.

Among the neural network approaches, both RNN-1 and RNN-2 demonstrated commendable
performance, indicating their efficacy in addressing the question answering task. The CNN
approach also exhibited competitive results, showcasing its ability to capture relevant features
for generating accurate responses. This means that these models have the advantage of capturing both short-term and long-term information, delivering consistent performance at a
reasonable training cost.

However, the Transformer approach, which is renowned for its attention mechanism and
parallelism processing capabilities, is generally superior to the RNN-based and CNN models.
At all four measures, the evaluation scores show that the Transformer model has the most
stability. This indicates that the Transformer architecture excels at capturing both short-range
and long-range dependencies of contextual information, underscoring its advantages for
question-answering tasks.

Significantly, GPT-2, a generative pre-trained model, emerged as the promising result in our
study. In terms of ROUGE-L score, it consistently outperformed all other approaches,
including the Transformer, RNN-1, RNN-2, and CNN. The robustness and language generation
capabilities of GPT-2 make it a highly effective model for generating accurate and contextually
appropriate answers.

On the BLUE-1, BLUE-4 and METEOR measures, the results are not good because of the
answer length generated by the model while these measures evaluate the accuracy of the
adjacent words. However, in general, the GPT-2 model still gives more natural results than the
previous ones.

The limitations of the Vietnamese GPT model in question answering can indeed be attributed
to its pre-training on data that does not specifically target this task. In particular, the pretrained
models of tuanle and danghuy1999 trained on news and wikipedia datasets, respectively, may
not have much information about the COVID-19 domain and have limitations on the amount
of training data. As a result, the model may not perform as well as the Transformer model in
certain scenarios. However, it is worth noting that the performance of the Vietnamese GPT
model, as measured by the ROUGE-L metric, exhibits better accuracy compared to other
evaluation metrics.
\begin{table*}[!h]

\centering
\caption{Comparison of GPT-2 models with other model.}
\begin{tabular}{llllllll}
\hline
\multicolumn{1}{l}{\textbf{Model}}                              & \multicolumn{1}{l}{\textbf{$n_{ans}$}} & \multicolumn{1}{l}{\textbf{BLEU-1}} & \multicolumn{1}{l} {\textbf{BLEU-4}} &  \multicolumn{1}{l}{\textbf{METEOR}}  & \multicolumn{1}{l}{\textbf{ROGUE-L}}\\ \hline
\multicolumn{6}{c}{\textbf{Traditional Neural Network}}                                                                                                    \\ \hline
\multirow{4}{*}{RNN-1} & \multicolumn{1}{l}{1}                                       & \multicolumn{1}{l}{6.54}                                 & \multicolumn{1}{l}{7.34}                            & \multicolumn{1}{l}{15.13} & \multicolumn{1}{l}{17.68} &                          \\
& \multicolumn{1}{l}{2}                                       & \multicolumn{1}{l}{6.85}                                 & \multicolumn{1}{l}{7.24}                            & \multicolumn{1}{l}{15.10} & \multicolumn{1}{l}{\textbf{18.01}} &   \\
& \multicolumn{1}{l}{3}                                       & \multicolumn{1}{l}{6.45}                                 & \multicolumn{1}{l}{\textbf{7.39}}                            & \multicolumn{1}{l}{14.46} & \multicolumn{1}{l}{17.86} &   \\ 
& \multicolumn{1}{l}{4}                                       & \multicolumn{1}{l}{\textbf{7.04}}                                 & \multicolumn{1}{l}{6.99}                            & \multicolumn{1}{l}{\textbf{15.28}} & \multicolumn{1}{l}{17.71} &   \\
\hline
\multirow{4}{*}{RNN-2} & \multicolumn{1}{l}{1}                                       & \multicolumn{1}{l}{\textbf{6.10}}                                 & \multicolumn{1}{l}{6.77}                            & \multicolumn{1}{l}{\textbf{13.80}} & \multicolumn{1}{l}{17.69} &                          \\
& \multicolumn{1}{l}{2}                                       & \multicolumn{1}{l}{5.88}                                 & \multicolumn{1}{l}{7.28}                            & \multicolumn{1}{l}{13.47} & \multicolumn{1}{l}{17.76} &   \\
& \multicolumn{1}{l}{3}                                       & \multicolumn{1}{l}{5.96}                                 & \multicolumn{1}{l}{\textbf{7.40}}                            & \multicolumn{1}{l}{13.69} & \multicolumn{1}{l}{\textbf{18.29}} &   \\
& \multicolumn{1}{l}{4}                                       & \multicolumn{1}{l}{5.48}                                 & \multicolumn{1}{l}{6.80}                            & \multicolumn{1}{l}{11.87} & \multicolumn{1}{l}{16.72} &   \\ \hline
\multirow{4}{*}{CNN} & \multicolumn{1}{l}{1}                                       & \multicolumn{1}{l}{11.88}                                 & \multicolumn{1}{l}{7.71}                            & \multicolumn{1}{l}{13.18} & \multicolumn{1}{l}{24.37} &                          \\
& \multicolumn{1}{l}{2}                                       & \multicolumn{1}{l}{\textbf{14.55}}                                 & \multicolumn{1}{l}{\textbf{9.11}}                            & \multicolumn{1}{l}{\textbf{13.21}} & \multicolumn{1}{l}{\textbf{25.97}} &   \\
& \multicolumn{1}{l}{3}                                       & \multicolumn{1}{l}{14.47}                                 & \multicolumn{1}{l}{8.84}                            & \multicolumn{1}{l}{12.44} & \multicolumn{1}{l}{25.23} &   \\
& \multicolumn{1}{l}{4}                                       & \multicolumn{1}{l}{13.36}                                 & \multicolumn{1}{l}{8.29}                            & \multicolumn{1}{l}{11.24} & \multicolumn{1}{l}{23.97} &   \\
\hline
\multicolumn{6}{c}{\textbf{Traditional Transformer}}    
 \\ \hline
\multirow{4}{*}{Transformer} & \multicolumn{1}{l}{1}                                       & \multicolumn{1}{l}{7.07}                                 & \multicolumn{1}{l}{7.07}                            & \multicolumn{1}{l}{10.19} & \multicolumn{1}{l}{16.65} &                          \\
& \multicolumn{1}{l}{2}                                       & \multicolumn{1}{l}{14.59}                                 & \multicolumn{1}{l}{9.55}                            & \multicolumn{1}{l}{13.08} & \multicolumn{1}{l}{22.52} &   \\
& \multicolumn{1}{l}{3}                                       & \multicolumn{1}{l}{\underline{\textbf{15.52}}}                                 & \multicolumn{1}{l}{9.07}                            & \multicolumn{1}{l}{\underline{\textbf{16.05}}} & \multicolumn{1}{l}{\textbf{28.59}} &   \\
& \multicolumn{1}{l}{4}                                       & \multicolumn{1}{l}{13.30}                                 & \multicolumn{1}{l}{\textbf{10.41}}                            & \multicolumn{1}{l}{12.47} & \multicolumn{1}{l}{22.85}
\\ \hline

\multicolumn{6}{c}{\textbf{Generative Pre-trained Transformer}}                                                                                                                  \\ \hline
\multirow{4}{*}{GPT-2} & \multicolumn{1}{l}{1}                                       & \multicolumn{1}{l}{0.81}                                 & \multicolumn{1}{l}{\textbf{9.47}}                            & \multicolumn{1}{l}{2.85} & \multicolumn{1}{l}{\underline{\textbf{29.05}}} &                          \\
& \multicolumn{1}{l}{2}                                       & \multicolumn{1}{l}{0.90}                                 & \multicolumn{1}{l}{8.92}                            & \multicolumn{1}{l}{3.22} & \multicolumn{1}{l}{28.60} &   \\
& \multicolumn{1}{l}{3}                                       & \multicolumn{1}{l}{\textbf{0.98}}                                 & \multicolumn{1}{l}{8.76}                            & \multicolumn{1}{l}{\textbf{3.42}} & \multicolumn{1}{l}{28.05} &   \\
& \multicolumn{1}{l}{4}                                       & \multicolumn{1}{l}{0.97}                                 & \multicolumn{1}{l}{8.54}                            & \multicolumn{1}{l}{3.32} & \multicolumn{1}{l}{27.82}  
\\ \hline
\multirow{4}{*}{GPT-2
(tuanle)} & \multicolumn{1}{l}{1}                                       & \multicolumn{1}{l}{0.73}                                 & \multicolumn{1}{l}{\textbf{8.41}}                            & \multicolumn{1}{l}{2.52} & \multicolumn{1}{l}{\textbf{24.82}} &                          \\
& \multicolumn{1}{l}{2}                                       & \multicolumn{1}{l}{0.88}                                 & \multicolumn{1}{l}{8.12}                            & \multicolumn{1}{l}{2.99} & \multicolumn{1}{l}{24.14} &   \\
& \multicolumn{1}{l}{3}                                       & \multicolumn{1}{l}{0.94}                                 & \multicolumn{1}{l}{8.12}                            & \multicolumn{1}{l}{3.18} & \multicolumn{1}{l}{24.30} &   \\
& \multicolumn{1}{l}{4}                                       & \multicolumn{1}{l}{\textbf{1.00}}                                 & \multicolumn{1}{l}{8.33}                            & \multicolumn{1}{l}{\textbf{3.30}} & \multicolumn{1}{l}{24.69}  
\\ \hline
\multirow{4}{*}{GPT-2
(danghuy1999)} & \multicolumn{1}{l}{1}                                       & \multicolumn{1}{l}{1.03}                                 & \multicolumn{1}{l}{\underline{\textbf{10.92}}}                           & \multicolumn{1}{l}{3.24} & \multicolumn{1}{l}{\textbf{23.41}} &                          \\
& \multicolumn{1}{l}{2}                                       & \multicolumn{1}{l}{1.18}                                 & \multicolumn{1}{l}{10.50}                            & \multicolumn{1}{l}{3.79} & \multicolumn{1}{l}{20.70} &   \\
& \multicolumn{1}{l}{3}                                       & \multicolumn{1}{l}{1.23}                                 & \multicolumn{1}{l}{10.38}                            & \multicolumn{1}{l}{3.95} & \multicolumn{1}{l}{21.01} &   \\
& \multicolumn{1}{l}{4}                                       & \multicolumn{1}{l}{\textbf{1.35}}                                 & \multicolumn{1}{l}{10.59}                            & \multicolumn{1}{l}{\textbf{4.19}} & \multicolumn{1}{l}{21.43}  
\\ \hline
\end{tabular}
\label{tab:ex}
\end{table*}
\section{Conclusion and Future Work}
In conclusion, this paper has explored the application of Generative Pre-trained Transformer
(GPT), a pre-trained language model, in the domain of natural language processing. Through
empirical studies, it has been demonstrated that GPT holds significant potential for question
answering systems, achieving impressive performance across various tasks. However, the
research on applying GPT specifically to the Vietnamese language is currently limited.
To address this gap, our study has presented an implementation of GPT for community-based
question answering, focusing on COVID-19 related queries in Vietnamese. We have
introduced a novel approach by conducting a comparative analysis between different
Transformer models and state-of-the-art (SOTA) models using a community-based COVID19 question answering dataset. The experimental results have revealed the highly promising
outcomes of the GPT-2 models, surpassing both SOTA models and previous community-based
COVID-19 question answering models developed for the Vietnamese language.
These findings highlight the potential and effectiveness of GPT-2 in addressing COVID-19
related queries within the Vietnamese community. The research contributes to expanding the
understanding of GPT's capabilities and its practical application in the Vietnamese language.

Future studies could further explore and refine the implementation of GPT for communitybased question answering in other domains and languages, promoting the development of
advanced question answering systems. In future research, we plan to extend our analysis to
include GPT-3 and ChatGPT, utilizing the OpenAI License Key. This will enable us to perform
a more comprehensive evaluation of these advanced language models within the context of community-based COVID-19 question answering in the Vietnamese language. By
incorporating these models into our study, we aim to explore their capabilities and assess their
performance in addressing the specific challenges and requirements of the Vietnamese
community.

\bibliography{sn-bibliography}

\end{document}